\documentclass[runningheads, hidelinks]{llncs}
\usepackage{graphicx}
\usepackage{cite}
\usepackage[ruled,vlined]{algorithm2e}
\usepackage{algorithmic}
\usepackage{booktabs}
\usepackage{url}
\usepackage[bookmarks=false]{hyperref}
\usepackage[font={footnotesize}, labelfont=bf]{caption}
\usepackage{bbding}
\begin{document}
	\title{Pattern Learning for Detecting Defect Reports and Improvement Requests in App Reviews} 
	\titlerunning{Pattern Learning for Detecting Defect Reports and Improvement Requests}
	%
	%
	\author{Gino V.H. Mangnoesing\inst{1} \and  Maria Mihaela Tru\c{s}c\v{a}\inst{2} \inst{(}\Envelope\inst{)} \and  \\
		Flavius Frasincar\inst{1}~\orcidID{0000-0002-8031-758X}}
	%
	%
	\institute{Erasmus University Rotterdam, Burgemeester Oudlaan 50,
		3062 PA Rotterdam, the Netherlands \and
		Bucharest University of Economic Studies, Piata Romana 6, 010374 Bucharest, Romania \\
		\email{gvh.sing@gmail.com},
		\email{maria.trusca@csie.ase.ro},
		\email{frasincar@ese.eur.nl}\\
	}
\maketitle              
\begin{abstract}
Online reviews are an important source of feedback for understanding customers. In this study, we follow novel approaches that target this absence of actionable insights by classifying reviews as defect reports and requests for improvement. Unlike traditional classification methods based on expert rules, we reduce the manual labour by employing a supervised system that is capable of learning lexico-semantic patterns through genetic programming. Additionally, we experiment with a distantly-supervised SVM that makes use of noisy labels generated by patterns. Using a real-world dataset of app reviews, we show that the automatically learned patterns outperform the manually created ones, to be generated. Also the distantly-supervised SVM models are not far behind the pattern-based solutions, showing the usefulness of this approach when the amount of annotated data is limited. 

\end{abstract}

\section {Introduction}

In the two last decades, the growth of user-generated content on the Web has accelerated enormously. This acceleration is bolstered by parallel developments, such as increased Internet access, increased Internet speeds, technological advancements in mobile devices, the growth of e-commerce and online payment methods, and many more. An important source of user-generated content with respect to customer feedback are online reviews, usually interpreted using Sentiment Analysis (SA) methods. The aim of SA is to automatically detect positive, neutral, and negative sentiments \cite{liu2015sentiment}.

A major downside of SA is that it measures satisfaction at a certain point in time. In this light, we argue that in addition to SA, it is important to focus on detecting specific types of feedback that indicate potential causes and influence factors of satisfaction. We consider such specific customer feedback as actionable, since it suggests a clear course of action for addressing the feedback, and thus directly help to modify and hopefully improve products.

In this paper, we focus on customer feedback related to mobile software applications which we will refer to as ``apps". We argue that software reviews are very important for aggregating valuable feedback. Firstly, because many companies have come to realise that all the technology required to transform industries through software is available on a global scale \cite{andreessen2011software}. Secondly, the field of software engineering has the well-accepted notions of bugs and feature requests, which we argue, are actionable types of feedback.

We approach feedback detection as a multi-label classification problem based on knowledge-base rules, in which our goal is to automatically determine if a given review is an example of actionable feedback. 
In general, managing and adapting a knowledge-based method is very tedious and labour intensive, especially because the considered domain that might have different (actionable) types of feedback. Therefore the developers of extraction rules or patterns need to have a proficient amount of domain knowledge, which might be rare. As a result, making a knowledge base of patterns is impractical to manage over time and across different domains. 
In this light, we suggest a system that is capable of performing pattern construction in an automated manner using genetic programming. Keeping in mind the importance of reduction of the human control over the system's design, we also tackle the problem of having a small number of labeled reviews (gold labels) using noisy labels generated based on patterns in a distantly-supervised way \cite{sahni2017efficient, ji2017distant}. The employed dataset and the proposed framework implemented in Scala are available at \url{https://github.com/mtrusca/PatternLearning}.


The remaining parts of the paper are structured as follows. In Sect. 2 we examine the most relevant works. Section 3 presents a detailed overview of the proposed framework in this study. In Sect 4 we evaluate our framework through a series of experiments. Finally, in Sect. 5 we present our conclusions and suggest future work.

\section{Related Works}
SA measures are not actionable, which means that further investigation is required to discover causes and opportunities for making improvements. While user-generated content is clearly an important source of customer feedback for businesses, only a few works are focused on exploring such data, beyond sentiment analysis.

There are very few works \cite{brun2013suggestion, goldberg2009may, beyondsentiment, ramanand2010wishful, qiao2017domain} that aim to detect specific information in customer feedback. The majority of these works only aggregate broad insights, mainly for the purpose of further exploration. In that sense, the information that is extracted through such approaches, can barely be considered actionable, as is also the case for the sentiment scores. 

Among the aforementioned works, only the method proposed in \cite{beyondsentiment} is more refined and suggests classifying actionable feedback, through a business-oriented approach. Namely, in \cite{beyondsentiment} lexical patterns are used to train a supervised classifier, rather than directly employing patterns for information extraction, which makes the extraction mechanism more adaptive to the various representations of feedback. Further on, this system summarizes the extracted feedback by means of a \textit{Topic Model}, generated using a clustering technique called \textit{Latent Dirichlet Allocation} (LDA) \cite{blei2003latent}. However, while the objective is very relevant, the suggested methods require a vast amount of manual labour to create useful feedback patterns. We argue this to be a great limitation since analysing customer feedback is an important process that should ideally be performed in a continuous fashion. Nevertheless, the study conveys a promising direction for future research in opinion mining, and clear feedback types to focus on, which we adopt in this work.
 
Pattern-based information extraction methods achieve machine comprehension of texts by transforming the unstructured texts to structured data. Some of the first methods that define pattern-matching engines to detect regular expressions in annotations on documents were proposed in \cite{cunningham1999jape, hearst1998automated}. Later on, in \cite{ijntema2012lexico} it was proposed a pattern language called \textit{Hermes Information Extraction Language} (HIEL), that creates information extraction patterns in a simpler (more compact) fashion compared to the aforementioned works, by including semantic information.

Usually pattern-based methods use diverse sets of rules that hone the process of information extraction. While the common rule learning methods like \textit{SEER} \cite{hanafi2017seer}, \textit{TANGO} \cite{jimenez2016learning}, or \textit{UIMA Ruta} \cite{kluegl2016uima} have proven to be effective for information extraction, they are not directly helpful for our goal to label documents that include a particular type of feedback. Since we are not sure which forms the feedback type might take, we require different patterns per target feedback. A more promising approach, would be to focus on discovering patterns, for which metaheuristics (e.g., evolutionary algorithms, simulated annealing, or particle swarm optimization) are more suited. As example, in \cite{borg2010automatic} a \textit{Genetic Programming} approach is employed to learn grammatical rules helpful in discriminating between definitions and non-definitions of terms. 
However, in \cite{castellanos2010leveraging} a \textit{Genetic Algorithm} \cite{booker1989classifier} is used to learn the most relevant combinations of prefixes and suffixes for an entity type of interest, extracted from the context they are in. This approach is similar to our work, since we also aggregate frequent combinations of words and tags, in the context of a certain feedback type, to include in patterns. However, we also include entity types as building blocks for generating patterns, which gives to our patterns extra flexibility.
   
\section {Methods}

In this research, our goal is to automatically detect actionable feedback in reviews. More specifically, we aim to detect two specific types of feedback: defect reports and improvement requests. We approach this task as a binary document classification problem, meaning that each review is considered a document that requires two classifications, one for each feedback type. Using this setup it is possible to classify some reviews as both defect report and improvement request.

In terms of functionality, a clear distinction can be made between two primary subsystems of our framework, namely the \textit{Prediction Engine}, and the \textit{Learning Engine}. Even though both subsystems share numerous services from the framework core module, their purpose is very different. In Sect. \ref{Prediction_Engine} we present the \textit{Prediction Engine}, and in Sect. \ref{Learnin_engine} we describe the \textit{Learning Engine}.

\subsection {Prediction Engine} \label{Prediction_Engine}

The \textit{Prediction Engine} is an essential system in our framework that exposes two methods, namely \textit{train} and \textit{predict}. The employed classifier is a linear SVM model, often applied to information retrieval problems. An important advantage for applying SVMs for learning text classification models is related to the fact that SVMs learning capability is independent of the dimensionality of the feature space. This property is highly beneficial given the complex nature of  natural language, which causes large and sparse features vectors in the featurization process.

\subsection {Learning Engine} \label{Learnin_engine}
Our main contribution to the research problem is to automate the task of discovering and constructing patterns. Rather then direct supervision, where labels are provided by human annotators, we use a group of patterns to provide (noisy) labels for each feedback type, which are then provided as input to the \textit{Prediction Engine}. Using noisy labels to guide algorithms, is a technique called \textit{Distant Learning} or \textit{Distant Supervision} \cite{sahni2017efficient, ji2017distant}. Despite the fact that \textit{Distant Supervision} is already a great step towards minimizing the amount of human labour required to perform feedback detection, the required process for manually constructing groups of patterns per feedback type, remains rather tedious, time consuming, and even requires specific knowledge about a pattern language, such as Regular Expressions, which might not be available for experts. For this reason, we suggest another level of automation, which is to automate the pattern creation procedure (responsible to generate noisy labels) by means of a learning algorithm.

To solve our problem for learning patterns, we require to select a learning algorithm that stands out with respect to interpretability and modifiability. The first requirement is important because the relationship between input and output has to be easily interpretable (for inspection). Secondly, we need an algorithm that allows us to provide the building blocks that form the search space of the problem, in order to be able to add features to avoid poor solutions, or even slow convergence. An example of building blocks is represented by the semantic features, which we need to add with the intention to increase coverage and to reduce ambiguity. A specific category of algorithms that meets these requirements are \textit{Evolutionary Algorithms} (EAs).

The most popular type of EA is the \textit{Genetic Algorithm} (GA), however we adopt a special case of GA called \textit{Genetic Programming} (GP) inspired by Darwin's theory of evolution \cite{booker1989classifier}. Genetic Programming and Genetic Algorithms are very similar. They both evolve solutions to a problem, by comparing the \textit{fitness} of each candidate solution in a population of potential candidates, over many generations. In each generation, new candidates are found by quasi-randomly changing (mutation) or swapping parts (crossover) of other candidates. The least ``fit" candidates are removed from the population. The primary difference between GA and GP is the representation of the candidate solutions. In GA a candidate is represented as a vector, and in GP a candidate is represented as a tree. As the GP representation fits better the specification of our information extraction patterns, we adopt it in our research.

The learning approach suggested in GP, is to define an environment in which a collection of randomly generated, simple programs (individuals) evolve through an analogue of natural selection. Each individual is composed from a predefined set of building blocks, which influence its so-called \textit{fitness} and thus its ability to survive in an environment with other individuals. 
In the GP architecture, individuals are represented by tree structures. Each tree is composed from a collection of nodes. All nodes (except the first, or root node) have one parent and any number of children. Every node belongs to one of two types, namely \textit{functions} or \textit{terminals}. Function nodes are allowed to have children nodes, which can be either functions or terminals. Terminal nodes are not allowed to have child nodes, therefore terminal nodes are considered the leaves of the tree. In our framework, we consider each individual to be a pattern for classifying documents (app reviews) with a (recursive) match method and propose a few different nodes for both functions and terminals, which are displayed in Table \ref {table:node_types}.

\begin{table}[t]
\centering
\caption{Overview of tree nodes}
\begin{tabular}{l l} 
 \hline
 Functions & Terminals \\ [0.5ex] 
 \hline
 AND Operator & Literal \\ 
 OR Operator & Part-of-Speech \\
 NOT Operator & Wildcard \\
 Sequence & Entity Type \\
 Repetition &   \\ [1ex] 
 \hline
\end{tabular}
\label{table:node_types}
\vspace{-0.5cm}
\end{table}

Function nodes include Boolean operations, such as \textit{AND}, \textit{OR}, and \textit{NOT}, as well as \textit{Sequence} and \textit{Repetition}. All functions are allowed to have an arbitrary number of children ranging from 1 and a predefined maximum. In some cases, a minimum number of children is required (e.g., 2 children for \textit{AND}). The \textit{Sequence} node can have one or more child nodes of types function or terminal. It is also the root node of each tree. A \textit{Repetition} node enforces two or more consecutive nodes to obey the same condition. A node of type \textit{AND} has at least two children, and is useful to pattern match for multiple features, for example to check whether a given token is both a specific literal and part of a syntactic category. The nodes of type \textit{OR} and of type \textit{NOT} also follow the Boolean logic, where the \textit{OR} nodes match as true if at least one of the children matches, and nodes of type \textit{NOT} match as true if none of its children match for a given token.

Terminal nodes are the external points (or leaves) of the tree. They are assigned a specific value, used to pattern match for specific tokens. \textit{Literal} nodes must be exactly matching the specific word (value) that is assigned the node. For \textit{Part-of-Speech} (POS) nodes, tokens are evaluated to match a specific Part-of-Speech tag. A \textit{Wildcard} node will match any token, irrespective of its value. Finally, an \textit{Entity Type} node matches a value from a manually constructed and populated gazetteer.  

Typically gazetteers consists of sets of terms containing names of entities such as cities, organisations, or weekdays \cite{cunningham2002gate}. Since at the time of performing this research, we could not find gazetteers for our specific domain, we decided to define our own. Our gazetteer is implemented using a plain key-value mapping, where a key corresponds to the name of an entity type, and the value stores a set lexical representations of that entity type. For example, to detect the entity type \textit{app} we employ the following terms: \textit{it}, \textit{app}, \textit{application}, \textit{Evernote} (we use a set of \textit{Evernote} reviews for our experiments). Some other entity types in our gazetteer are: \textit{user}, \textit{action}, \textit{object}, \textit{component}, \textit{device}, and \textit{update}. The entity types we employ are inspired by \textit{Issue Tracking Systems} (ITS), such as \textit{Bugzilla}, an open-source issue tracker created by \textit{Mozilla}. Since ITS involve very comparable types of feedback to this study, we consider the entity types in ITS a useful starting point for constructing our gazetteers. 

The first step for each genetic program, is to generate an initial population of $N$ individuals. Although, this step seems very straightforward conceptually, it is slightly more complex in practice. The fundamental purpose of the initialization phase, is to generate a population with the ``potential" to evolve and produce strong offspring, through natural selection and genetic operations. In this light, we aim to include the most promising configuration, functions and terminals.

There are a range of methods to generate the initial population. The two most popular
methods, are \textit{full} and \textit{grow}. The \textit{full} method generates trees in such a way that all branches grow until reaching their limit. While the tree develops, starting from the root node, nodes are generated at random and assigned as child nodes. Only function nodes are considered when generating child nodes before the maximum depth is achieved, after this point only terminals nodes being produced. As a result, all the terminal nodes are at the same level of depth in the tree, which is the maximum depth. The \textit{grow} method develops in a similar manner, by randomly assigning child nodes to extend into branches. However, each child node is selected from both function and terminal nodes, which results in branches of various shapes and lengths, only restricted by the maximum depth. 

In our experiments, we use the \textit{ramped-half-and-half} method \cite{koza1992genetic}, which is commonly used since it produces a wider range of variation in terms of shapes and sizes of trees compared to the \textit{grow} or \textit{full} methods, hence, increasing the likelihood of including fit individuals in the population. The \textit{ramped-half-and-half} achieves more variety, by combining both the \textit{grow} and \textit{full} methods for generating individuals, where one half of the population is generated through the \textit{grow} method, and the other half through the \textit{full} method. The algorithm we employ to generate individual trees in a recursive manner is based on the one suggested in \cite{ijntema2014genetic}.








During the initialization, nodes are selected randomly to construct trees. However, for the purpose of stimulating useful combinations of terminals, we generate a pool of recommended terminal candidates. Whereas the pool contains all entity types and the wildcard, for the case of \textit{Part-Of-Speech} (POS) and \textit{Literal} terminals we select only the most relevant nodes. More specifically, we pre-analyse the training set for frequently occurring unigrams and bigrams of types \textit{Literal} and \textit{Part-of-Speech} (POS). For bigrams, four specific pair combinations are considered, namely: (\textit{Literal})(\textit{Literal}), (\textit{POS})(\textit{Literal}), (\textit{Literal})(\textit{POS}), and (\textit{POS})(\textit{POS}). Subsequently, we remove in each sentiment class of a target feedback type, the 100 most frequent unigrams and bigrams that occur in the another sentiment class. Then, every time a terminal node is needed we randomly selected from the pool of recommended terminal candidates. 

An important measure of quality for individuals is captured by the so-called \textit{fitness} measure. The \textit{fitness} of our individuals, i.e., patterns, can be determined in a manner that is common for Information Retrieval methods, namely through calculating $F_1$-measure. However, in our problem, we want individual patterns to be optimized for high precision, which means that we want more weight on precision than recall. Hence, we have to employ the $F_\beta$-measure with $\beta=0.3$ (recall is assured by learning multiple patterns for a specific feedback type forming a pattern group).

In Evolutionary learning methods, a population of individuals can evolve for many generations. However, after a certain amount of generations, the \textit{fitness} of the best new individuals will either stop increasing. In our framework, we employ two criteria for termination. The first criteria is the maximum number of generations and is checked when generating a pattern (in the pattern group). The second criteria is checked per event type and it is triggered if the the pattern does not increase the \textit{fitness} of the entire group of patterns after a maximum number of iterations. The \textit{fitness} measure for a group of patterns is determined by the $F_1$-measure, instead of the $F_\beta$-measure. Our motivation for using $F_1$ for group fitness is related to our goal to seek patterns for as many variations of a target feedback type as possible.


A proper procedure for selection should not find only the strongest individual of a population, but to allow more individuals to have a chance of being selected. A common method that addresses this requirement is \textit{Tournament Selection}. Precisely, the method allows for a constant selection pressure that determines the extent to which fit individuals are preferred over less fit individuals. All the selected individuals are used to produce \textit{offspring} or the next generation of individuals. The main objective in producing offspring, is to enhance the \textit{fitness} for the next generation based on three genetic operations, namely \textit{Elitism}, \textit{Crossover}, and \textit{Mutation}. 

As discussed earlier, our goal is to learn a group of patterns that detect as many variations of a target feedback type as possible, in our training examples. In essence, each pattern can be interpreted as a rule, and each document has to be categorised as either positive or negative, according to our ``knowledge" of each category, which is stored in a rule base. The set of rules learnt in our framework is generated through a \textit{Sequential Covering Algorithm}, that complements our second GP termination criteria.

\section{Experiments}

In order to evaluate the approach suggested in our framework, we performed experiments on a real-life dataset. The dataset contains 4470 reviews about \textit{Evernote}, a mobile app for the Android platform. We automatically extracted the review dataset from the Google Play Store, through Web scraping techniques. We selected \textit{Evernote} because it is a widely used app with a large user base, that publicly share their feedback on the Web, and therefore serves as a great example for our examined research problem.

\begin{figure}[t]
\centering
  \includegraphics[width=8cm]{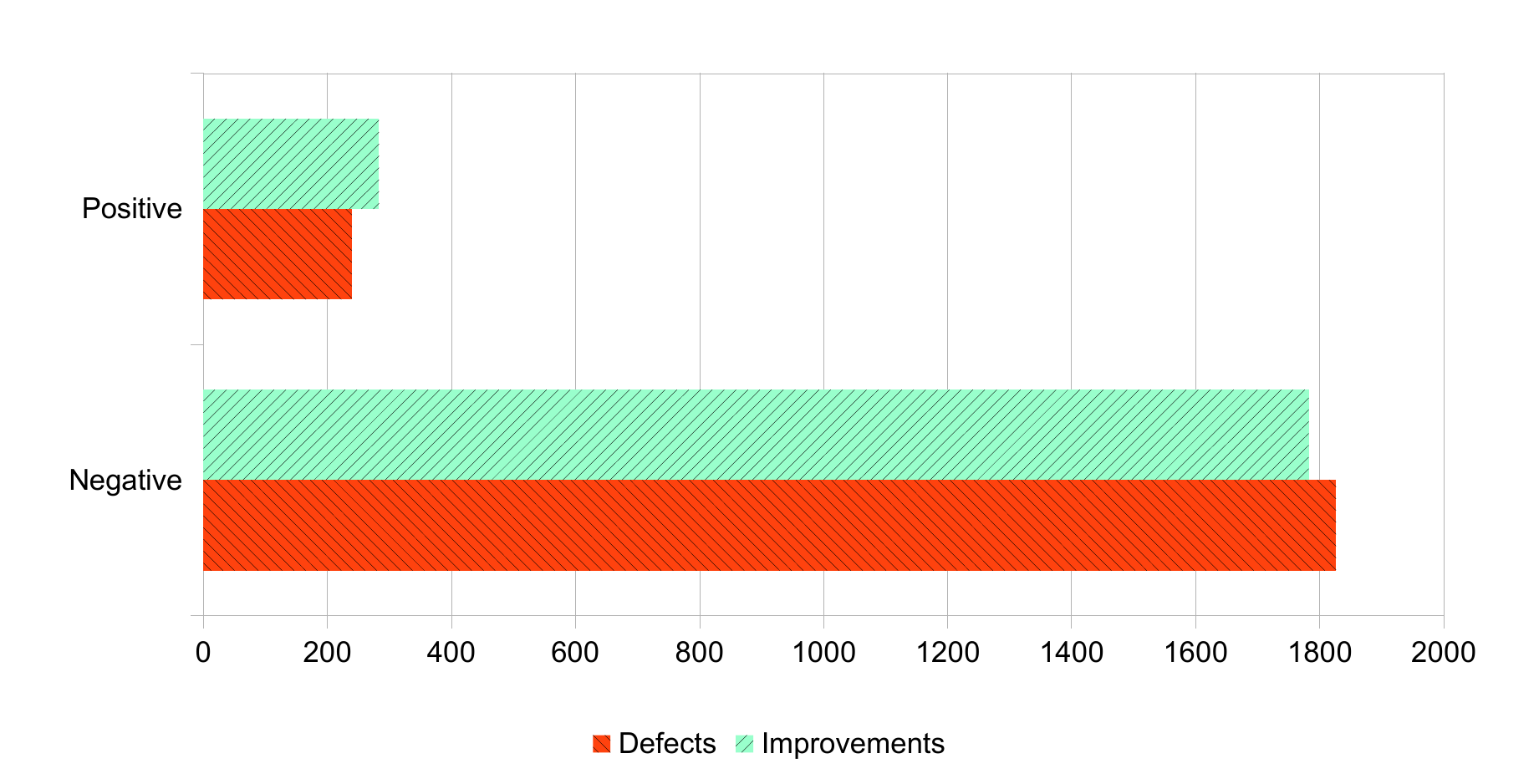} \\
  \caption{Data distribution per feedback type}
  \label{fig:feedback_distribution}
\end{figure}

\begin{table}[t]
	\centering
	\caption{Inter-annotator agreement per task}
	\begin{tabular}{ lc } 
		\hline
		Question & Agreement(\%)\\
		\hline 
		Does this review contain a defect report? & 97.1 \\ 
		Does this review contain an improvement request? & 92.75 \\ 
		\hline
	\end{tabular}
	\label{table:annotator_agreement}
\end{table}

We have annotations for 46\% of the total review dataset. We hold out 20\% of all reviews for testing purposes in all methods. Therefore, we have the remaining 26\% of reviews available for training purposes. However, for the experiments that employ distant supervision, we generate noisy labels, hence, have 80\% of the full review dataset available for training. In Fig. \ref{fig:feedback_distribution} we depict the distribution of both defects and improvements in the labeled collection of reviews. The terms ``Positive" and ``Negative" refer to the classification labels that were assigned to every review per feedback type by human annotators. On average 12.6\% of our labeled set of reviews contains one or more actionable types of feedback, in which there are 8.4\% more requests for improvement than defect reports. Finally, only 1.3\% of our annotated reviews is labeled as both a defect report and an improvement request.

We collected annotations for both feedback types through \textit{CrowdFlower} (recently renamed \textit{Figure Eight}), an online data enrichment platform. The instructed task is to label every individual review for both defect reports and improvement requests. Every review was annotated by at least 3 annotators, and in some cases even 5 or 7 (when it is recorded a low accuracy of the test questions that inspect the quality of the annotator). In order to estimate the reliability of the aggregated annotations, we computed the inter-annotator agreement (IAA) for each question, using \textit{Fleiss' Kappa} measure. As can be observed in Table \ref{table:annotator_agreement}, the inter-annotator agreement for both questions is very high, which indicates that different contributors performed similar judgements consistently. We use the majority voting for building our gold standard.

\begin{table}[t]
\footnotesize
\centering
\setlength{\tabcolsep}{0pt} 
\caption{\footnotesize {Examples of human (A) and automatically constructed (B) patterns. DR and IR stand for Defect Report and Improvement Request, respectively. For DR patterns ``:" separates the terminal from its type.}}
\resizebox{\textwidth}{!}{%
\begin{tabular*}{\textwidth}{@{\extracolsep{\fill}\quad}lll}
\toprule
 Type & Pattern & Example\\
 \midrule
 A (DR) & i can (n't$\vert$not) & I can't remove the numbers in lists anymore.\\
 \cmidrule{2-3}
& no (option$\vert$ability) to & No ability to copy or duplicate notes on mobile. \\
\midrule
A (IR) & please VB & Please add Google now integration.\\
\cmidrule{2-3}
& (an$\vert$the) option to & Would like to see an option to adjust the font \\
& & size.\\
\midrule
B (DR) & OR: & However I cannot do so from the app which is \\
& $\vert$-(however$\vert$but): Literal & very appalling.\\
& $\vert$-(not$\vert$n't): Literal& \\
\cmidrule{2-3}
& OR: & The last few months of updates haven't changed \\
& $\vert$-Software Bug: Entity & or lessened the lag you get when you edit notes.\\
& Type &\\
& $\vert$-Software Update: Entity & \\
& Type & \\
\midrule
B (IR) & SEQ: & Please add automatic title from the  
rst sentence \\
& $\vert$-please: Literal & from notes instead of adding auto events...\\
& $\vert$-VB: Syn. Category& \\
\cmidrule{2-3}
& SEQ: & Colour coding of the notes and reminders for \\
& $\vert$-5: Literal & repetitive tasks can fetch 5 stars.\\
& $\vert$-stars: Literal & \\
\bottomrule
\end{tabular*}}
\label{table:table_3}
\vspace{-0.5cm}
\end{table}

The employed patterns are constructed both manually and automatically. In the \textit{Evernote} dataset, there are proposed five manual and two generated patterns for defects, and eight manual and ten generated patterns for improvements. The most likely reason for this contrast is the variation in distribution of feedback types in our dataset, as a result of the fact that \textit{Evernote} is a popular app, well tested, and optimised. Furthermore, we noticed that the most effective patterns only use function nodes of type \textit{Sequence} and \textit{OR}. Also, many examples of feedback can be recognized with a single terminal, such as the \textit{Entity Type} ``software update" for defect reports or the \textit{Literal} ``stars" for improvement requests, which indicates that the level of specificity does not necessarily have to be high. In that light, patterns that include the \textit{NOT} node, which requires feedback examples in which a very specific word is not mentioned are often not necessary. While \textit{NOT} functions can be useful to make a pattern very expressive and precise, it becomes obsolete when that level of selectivity is not required, as in our case. A similar line of reasoning can be applied to the \textit{AND} functions. Table \ref{table:table_3} lists two example of patterns for each pair (feedback type, pattern type).

\begin{table}[t]
\footnotesize
\centering
\setlength{\tabcolsep}{0pt} 
\caption{\footnotesize {Performance metrics for feedback type classifications in terms of precision, recall, and $F_1$-measure.}}
\resizebox{\textwidth}{!}{%
\begin{tabular*}{\textwidth}{@{\extracolsep{\fill}\quad}lcccccc} 
\toprule
\multicolumn{1}{c}{Task}
& \multicolumn{3}{c}{Defect Classification}
& \multicolumn{3}{c}{Improvement Classification} \\
\midrule
Method & Precision & Recall & $F_1$-measure & Precision & Recall & $F_1$-measure \\
\midrule
Standard SVM                & 0.39           & 0.59       & 0.47         & 0.78            & 0.54         & 0.64           \\ 
Patterns A (manual)         & 0.61           & 0.42       & 0.50         & 0.81            & 0.42         & 0.56           \\ 
Patterns B (learned)        & 0.91           & 0.39       & 0.54         & 0.79            & 0.51         & 0.62           \\ 
SVM Distant Supervision A   & 0.24           & 0.67       & 0.36         & 0.39            & 0.48         & 0.43           \\ 
SVM Distant Supervision B   & 0.41           & 0.59       & 0.49         & 0.46            & 0.44         & 0.45           \\ 
\bottomrule
\end{tabular*}}
\label{table:table_4}
\end{table}

\begin{table}[t]
\footnotesize
\centering
\setlength{\tabcolsep}{0pt} 
\caption{\footnotesize {Running time for pattern creation per approach.}}
\resizebox{\textwidth}{!}{%
\begin{tabular*}{\textwidth}{@{\extracolsep{\fill}\quad}lccc} 
\toprule
\multicolumn{1}{c}{Approach}
& \multicolumn{1}{c}{Defect Patterns}
& \multicolumn{1}{c}{Improvement Patterns}
& \multicolumn{1}{c}{Total}\\
\midrule
Manual (per person) & 8.5 hours       & 10.25 hours          & 18.75 hours \\ 
Automated & 3.5 hours       & 2.4 hours          & 5.9 hours   \\ 
\bottomrule
\end{tabular*}}
\label{table:table_5}
\vspace{-0.2cm}
\end{table}

To classify defect reports and improvement requests we test the following methods:

\vspace{1mm}
\noindent{\textbf{Method 0: Standard SVM.} In this method, we train an SVM classifier using only labelled reviews for training. This method can be considered a reference for the following methods.}

\vspace{1mm}
\noindent{\textbf{Method 1: Patterns A. } In this experiment, we use human patterns to perform supervised classifications directly (without SVMs). We employ the available labelled data (26\%) for learning patterns. }

\vspace{1mm}
\noindent{\textbf{Method 2: Patterns B. } This method is similar to the Method 1, except that the human patterns are replaced with automatically constructed ones.}

\vspace{1mm}
\noindent{\textbf{Method 3: SVM Distant Supervision A. } In this method, we train an SVM classifier using noisy labels generated based on the human patterns for the entire training set. }

\vspace{1mm}
\noindent{\textbf{Method 4: SVM Distant Supervision B. } This method is similar to the Method 3, except that the human patterns are replaced with automatically constructed ones.}

\vspace{1mm}


Table \ref{table:table_4} displays an overview of performance measures of all proposed methods. We can notice that the \textit{Distant Supervision} methods are not far behind the direct classification through patterns, in terms of $F_1$-scores. Nevertheless, given that the results are obtained with noisy labels shows the usefulness of this approach for datasets where the annotated data is limited. 

As regards, the comparison between the two types of patterns, it is obvious that the automatically generated patterns perform better than the human ones. In order to have a complete insight over the pattern creation process (manual versus automated) we additionally explore the patterns' efficiency besides their effectiveness. Table \ref{table:table_5} displays the running time for creating patterns both manually and automatically. We can observe that it takes 70\% less time to generate the automatic patterns than the manual ones. 

\section {Conclusion}
In this study we presented a framework for automatically learning lexico-semantic patterns helpful for detecting specific types of feedback expressed in conversational customer feedback (defect reports and improvement requests). Using a custom dataset, we showed that the automatically generated patterns perform slightly better than the manual ones and there is a 70\% reduction in construction time. Further on, we demonstrated that the distantly-supervised SVM with noisy labels is not far behind the pattern-based classification. The results reveals the applicability of this approach when the amount of available labels is limited. 

As our future work, we would like to increase the flexibility of our patterns by considering more complex terminal structures. Using techniques from entity-learning we would like to explore the automatic generation of our domain-specific gazetteers lists to increase coverage and the framework applicability in other domains.

\bibliographystyle{splncs04}
\bibliography{main}

\end{document}